\documentclass{article}

\usepackage{arxiv}

\usepackage[utf8]{inputenc} 
\usepackage[T1]{fontenc}    
\usepackage{hyperref}       
\usepackage{url}            
\usepackage{booktabs}       
\usepackage{amsfonts}       
\usepackage{nicefrac}       
\usepackage{microtype}      
\usepackage{cleveref}       
\usepackage{lipsum}         
\usepackage{graphicx}
\usepackage{natbib}
\usepackage{doi}
\usepackage{xcolor}
\usepackage{listings}
\usepackage[caption=false]{subfig}
\usepackage{enumerate}

\title{Wavelet leader based formalism to compute multifractal features for classifying lung nodules in X-ray images}

\date{}

\author{I.  M.  Sierra-Ponce,  A.  M. León-Mecías, D.  Valdés-Santiago \\\\ Applied Mathematics Department, Faculty of Mathematics and Computer Science \\ University of Havana \\ Cuba}

\renewcommand{\headeright}{}
\renewcommand{\undertitle}{}
\renewcommand{\shorttitle}{Wavelet Leaders Multifractal Formalism for nodule detection in chest X-rays}

\hypersetup{
pdftitle={Wavelet leader based formalism to compute multifractal features for classifying lung nodules in X-ray images},
pdfsubject={cs.MS},
pdfauthor={Isabella María Sierra Ponce, Ángela León Mecías,  Damian Valdés Santiago},
pdfkeywords={	images processing, lung nodules classification, support vector machine, multifractal features, wavelet-leader multifractal formalism},
}

\begin{document}
\maketitle

\begin{abstract}
This paper presents and validates a novel lung nodule classification algorithm that uses multifractal features found in X-ray images. The proposed method includes a pre-processing step where two enhancement techniques are applied: histogram equalization and a combination of wavelet decomposition and morphological operations. As a novelty, multifractal features using wavelet leader based formalism are used with Support Vector Machine classifier; other classical texture features were also included. Best results were obtained when using multifractal features in combination with classical texture features, with a maximum ROC AUC of 75\%. The results show improvements when using data augmentation technique, and parameter optimization. The proposed method proved to be more efficient and accurate than Modulus Maxima Wavelet Formalism in both computational cost and accuracy when compared in a similar experimental set up. 
\end{abstract}


\section{Introduction}
Lung cancer is a severe respiratory disease, with high prevalence in human beings and the highest oncological mortality rate around the globe. Achieving early diagnosis is critical for treatment: early stage lung cancer can be surgically treated in approximately 20\% of all cases. Chest X-ray is one of the most widely used lung cancer diagnostic tool. This technique shows a well-defined view of the heart and large blood vessels, as well as lung masses. It usually involves two complementary films: posteroanterior (70\% of lung area visibility) and lateral (30\% of lung area visibility). In a simple chest X-ray one can visualize abnormalities in 98\% of patients with bronchogenic carcinoma and 75\% of lung cancer variations can be detected, and between a 5\% and a 15\% of asymptomatic cases, \citep{kelly2012chest}.

Single pulmonary nodule usually appears in 33\% of all lung cancer cases and it is critical in differential diagnosis. Morphologically, it is rounded and densely shaped, and those features are related to image texture descriptors. In this work, multifractal analysis technique for feature extraction is proposed and tested in lung nodule detection task.   

Multifractal analysis is a novel technique in medical image processing: it studies the local regularity and scale behavior of functions; it is an attempt to describe geometrical and statistical distributions of the singularities of functions \citep{seuret2016multifractal}; therefore, multifractal analysis is also useful to describe images and its results have been satisfactory in segmentation and classification tasks. The multifractal spectrum is a function that measures local regularity of a signal, from a global point of view. This means that it studies the incidence the certain singularity exponents found in a structure. To estimate the multifractal spectrum a so called formalism is used.

\citep{kido1995fractal} performed fractal dimension computation on linear opacities that were previously extracted from chest X-rays, in order to study interstitial abnormalities in lungs. The fractal dimensions obtained from the regions of interest (ROIs) in lungs with interstitial abnormalities were significantly higher compared with those from ROIs in normal lungs, showing fractal features are relevant to discern between normal and abnormal lungs. \citep{rodriguez2005medidas} used fractal analysis to make a mathematical characterization of geometric structures present in chest X-rays of healthy patients.

\citep{silvetti2010analisis} used multifractal analysis for digital mammography and CT segmentation. Its application relies on \textit{box counting} technique, a computationally efficient alternative to fractal dimension computation. In the latter, multifractal analysis was applied to segmentation according the nature of tissue. In \citep{braverman2013scale}, a more generic family of functions of which \textit{box counting} is a special case is used: R$\grave{e}$nyi entropies of order $\alpha$. A study on breast tissue is carried out, relying on multifractal features for Kruskal-Wallis statistical analysis. It is stated that there are significant multifractal spectrum differences between at least two of the three tumor grades considered. 

\citep{maceda2014computo} proposed lacunarity calculus (it is a fractal feature to distinguish a structure's ability to fill the space that contains it) is proposed to study MRI brain images. In general, lacunarity is not conclusive on its own: features obtained are not significantly different within the sample set. 

In \citep{leonarduzzi2014analisis} \textit{p}-leader wavelet-based multifractal formalism is applied on a fetal heart rate database. Log-cumulants are used as descriptors, and were validated as an acceptable feature to distinguish between normal and pathological samples with a ROC AUC measure going from $\approx$ 60\% up to $\approx$ 73\%. 

In \citep{hernandez2015analisis} \textit{box counting} dimension was used, as well as other fractal features obtained using \lstinline|MATLAB| tool for evaluating retina images. In \citep{troshin2015multifractal} multifractal spectrum was used to parameterize regions of interest within a chest X-ray and then so detect asymmetrical lesions. 

\citep{marusina2017mri} examined healthy and abnormal liver MRI images through fractal dimension obtained using box-counting, and other self-similarity measure called Hurst exponent.

In \citep{lorenaleon} log-cumulants are used as multifractal feature, based on wavelet leader multifractal formalism, to classify malignity of masses in digital mammography. An average precision of 80\% was obtained, using K-Nearest Neighbors and Support Vector Machine.

\citep{lee2020automatic} segmented lung nodules in CT scans by using multifractal analysis. Method proves to be competitive among other state-of-the-art techniques, such as the ones proposed in \citep{xu2002automated} and \citep{kostis2003three}, since it is a fully automatic segmentation that does not depend on seed points.

In this work, a multifractal analysis method based on wavelet leaders as multiresolution quantity is proposed for lung nodule detection in chest X-rays. The paper is structured as follows: in section 2 some theoretical foundations about fractal and multifractal analysis are presented. The proposed methodology and some experiments are described in section 3. Section 4 is devoted to discussion of the main results. Finally some conclusions are given.

\section{Fractals and Multifractality}\label{section_2}
	Fractal are complex mathematical objects that do not have a minimum natural measure. It is Mandelbrot \citep{mandelbrot1982fractal} who first noted the relevance of fractals in many fields such as physics, while making contributions with numerical and experimental studies. The concept of Hausdorff dimension is the core of fractal geometry, but the calculations involved in its computation are mathematically complex; therefore, not feasible. In practice, alternatives such as box dimension \citep{LIEBOVITCH1989386} are used to estimate it. Fractal dimension is often defined as an exponent that describes how the structure of an object is repeated at different scales. When a structure has multiple fractal dimension values across the scales, it is called a \textit{multifractal}.

	\subsection{Multifractal Spectrum} \label{multifractal_spectrum}
	Multifractal spectrum \citep{multifractalspectrum}, denoted by $D$,  is a function that represents a quantitative description of the fractal nature of a structure, for example, signals like images. It is theoretically defined as the correspondence between the H\"older exponents (denoted by $h$) in \textit{E} (the image in this case), \citep{article}, and \textit{Hausdorff} dimension of related iso-H\"older sets. A set is called to be iso-H\"older if $\forall x \in E_{\alpha}, h(x)={\alpha}, \alpha \in \mathbb{R}$. When computing Hausdorff dimension we consider all possible coverings of $E$, to then choose the optimum and study its variation across the scales. 
Multifractal spectrum gives global information about local singularity distribution in a signal $X$. If $D(h)=0$, singularities linked to $h$ are isolated; and, if $D(h)=1$, $h$ is constant in an interval. 

	\subsection{Multiresolution Paradigm}

X-ray images have structural features that appear through different scales. This statement has brought to attention a new paradigm to study such images and similar ones: it is based on multiresolution quantities $T_X(j, x)$ as main descriptors for an image $X$ observed at a scale $j$ and at a specific point $x$. This approach makes it possible to study the behavior of a signal $X$ across a set of scales.  
Scale invariant paradigm can be described as the following decay rate in a \textit{power law}:

\begin{equation}
	\frac{1}{n_j}\displaystyle\sum_{k=1}^{n_j}|T_X(j, x)|^{\alpha} \sim c_qj^{\zeta (q)}, \;\;\; j \in \left[ j_{min}, j_{max}\right] 
\end{equation}
As a result, analysis now focuses in the $\zeta(q)$ exponents that describe the decay of $T_X(j, x)$ statistical moments.

The main advantage of multiresolution paradigm is that it gives us the ability to study non stationary phenomena and to describe samples through statistical moments of a higher order. 

\subsection{Wavelet Leader based Multifractal Formalism}

In order to find an adequate multifractal formalism to estimate multifractal spectrum of a function $X$, we need to define a multiresolution quantity $T_X(j, x)$ that varies across the scales $j$ according to a power law:

\begin{equation}\label{decay_power_law}
	\frac{1}{n_j}\displaystyle\sum_{k=1}^{n_j}|T_X(j, x)|^{\alpha} \sim c_qj^{\zeta (q)} \;\;\; j \in \left[ j_{min}, j_{max}\right] 
\end{equation}
with $\alpha_{x_0} = h_X(x_0)$, H\"older exponent in $x_0$. In practice, an estimator of statistical moments $q$ of $T_X(j, x)$ is used. 
A statistical moment of order $q$ of a discrete function $X$ with $K$ samples is defined by the equation:
	\begin{equation}\label{statistical_moment}
		S_X(q) = \frac{1}{K}\displaystyle \sum_{k=1}^{K} X(x_k)^q.
	\end{equation}
 A statistical moment describes the shape of a function. For example, if it is related to a probability distribution, the first statistical moment is the expected value and the second moment is the variance.

Let $S_T(q, j)$ be the statistical moment of order $q$ of the multiresolution quantity $T_X(j, x)$

\begin{equation}\label{estimator}
	S_T(q, j) = \frac{1}{n_j}\displaystyle \sum_{k=1}^{n_j}T_X(j, x_k)^q,
\end{equation}
with $n_j$ being the cardinality of the sample related to scale $j$.

If we consider the hypothesis of $T_X(j, x)$ varying according to a power law, then the following relation is true:

\begin{equation}\label{relation}
	S_T(q, j) \sim j^{\zeta_T(q)},
\end{equation}
and $\zeta_T(q)$ is the scaling function and it is related to the regularity of $T$ in the scale $j$. In practice, we can find $\zeta_T(q)$ through the following linear regression:
\begin{equation}\label{linear_regression}
	\zeta_T(q) \sim \frac{\log{S_T(q, j)}}{\log{j}}.
\end{equation}

We will build the notion of multifractal formalism based on an heuristic presented in \citep{leonarduzzi2014analisis} as follows:

\begin{enumerate}
	\item Let us assume that function $X \in \mathbb{R}^d$. We know that $n_j$ decreases as $j$ increases, and it does it by following a rate of $j^d$. Then, we can say that $\frac{1}{n_j} \sim j^d$.
	\item Let us take as hypothesis that multiresolution quantity $T_X(j, x)$ varies following a power law related to H\"older exponent $h$, so $T_X(j, x) \sim j^h$.
	\item The relation between the cardinality of the set of points that have the same H\"older exponent and the scale $j$ is defined by the power law $n_j \sim j^{-D(h)}$. 
\end{enumerate}

According to the latter in conjunction with equation (\ref{estimator}), we obtain:

\begin{equation}
	S_T(q, j) \sim j^d \cdot j^{qh} \cdot j^{-D(h)},
\end{equation}
and 
\begin{equation}
	\zeta_T(q) \sim d + qh - D(h).
\end{equation}

For scaling levels approaching 0, first order moments take more relevance:
\begin{equation}
	\zeta_T \sim inf_q(d+qh - D(h)),
\end{equation}
which leads to a \textit{Legendre} transform. If we consider inverse transform, let us have:

\begin{equation}\label{legendre_inverse}
	\mathcal{L}_T \sim inf_h(d+qh-\zeta(q)).
\end{equation}

If the function is convex, then $\mathcal{L}(h)= D(h)$. Otherwise, $\mathcal{L}$ is the convex hull of multifractal spectrum and $\mathcal{L}(h)\geq D(h)$. 

According to \citep{leonarduzzi2014analisis}, an adequate multiresolution quantity is such that varies according to power law of exponent $h$; also, it should be a hierarchical quantity and, if possible, be robust to smoothness. Robustness condition refers to the ability of multiresolution quantity $T_x$ to ignore certain grades of smoothness of a function. In this work, that concept is linked to vanishing moments of the mother wavelet, denoted $N_{\psi}$, which we will describe in more detail.     

A hierarchical function is that which satisfies:
	\begin{equation}
		A \subset B \Rightarrow f(A) \leq f(B).
		\label{hierarchical}
	\end{equation}
In the case of $T_X(j, x)$, the concept of subset is related to scales. Translating, we have:

\begin{equation}
	j<j^{\prime} \Rightarrow T_X(j, x) \leq T_X(j^{\prime}, x). 
\end{equation}

In this work, wavelet leader based multifractal formalism is used, as it effectively solves this issue \citep{wendt2009wavelet} . Details will be given in the following sections. 

\subsubsection{Wavelet Leader Coefficient Decomposition}
Let the bidimensional dyadic intervals be defined as:
\begin{equation}
	\lambda_{j, k_1, k_2} = \left\lbrace \left[k_12^j, (k_1 + 1)2^j \right] \times \left[ k_22^j, (k_2 +1)2^j \right]  \right\rbrace, 
\end{equation}
and the centered neighborhoods union set:

\begin{equation}
	3^2\lambda_{j, k_1, k_2} = \displaystyle \bigcup_{n_1, n_2 \in \{-1, 0, 1\}} \lambda_{j, k_1 +n_1, k_2 +n_2},
\end{equation}
the wavelet leader coefficients of a function $I \in L^2(\mathbb{R}^2)$ are defined as:
\begin{equation}\label{leaders}
	L_X(j, k_1, k_2) = \sup_{m=1, 2, 3, \lambda^{\prime} \subset 3^2 \lambda_{j, k_1, k_2}} |c_X^{(m)} (\lambda^{\prime})|,
\end{equation}
where $c_X^{(m)}(\lambda ^{\prime})$ represents two-dimensional wavelet coefficients $c_{X}^{(m)}(j',k'_1,k'_2),\; m=1,2,3$ \citep{mallat1999wavelet}. In practice, maximum coefficients in defined neighborhood, within scales $2^{j^{\prime}}\leq 2^j$ and $m$ directions are taken, \citep{wendt2009wavelet}.

In order to define wavelet leader based multifractal formalism, it is considered $T_X(j, x)= L_X(j, k_1, k_2)$ and the structure function $S_L(q, j)$ is defined as:

\begin{equation}
	S_L(q, j) = \frac{1}{n_j}\displaystyle \sum_{\lambda \in \Lambda_j} L_{X_{\lambda}}^q,
\end{equation} 	
and, as before the scaling function $\zeta_L(q)$ is defined based on $S_L(q, j)$:

\begin{equation}
	\zeta_L(q) = \liminf_{j \rightarrow - \infty} {\frac{\log{S_L(q, j)}}{\log{2^j}}}.
\end{equation} 

Wavelet leader based Legendre spectrum is defined in \citep{wendt2009wavelet} as follows: 

\begin{equation}
	\mathcal{L}_L = \inf{(1+hq+\zeta_L(q))}.
\end{equation}

\subsubsection{Practical Computation of Multifractal Spectrum}\label{practical_mfa}

Computation of scaling function $\hat{\zeta}(q)$ is done through linear regressions of $\log_2{S(q, j)}$ (base 2 logarithm is used for convenience according to wavelet analysis dyadic intervals) vs. $j$. 

Scales are critical parameters in multifractal analysis. Although the selection of scales can be difficult in some cases, a simple way of doing it is by the observation of graph $j$ \textit{versus} $\log_2S_L(q, j)$, searching for an interval $j \in \left[j_{min} ,j_{max} \right]$ in which the function has linear behavior. 

In practice, it is difficult to work straight with the functions $\zeta(q)$ and $\mathcal{L(h)}$, that is why in \citep{castaing1993log} $q=0$ centered Taylor series of function $\zeta(q)$ are used:

\begin{equation}
	\zeta(q)=\sum_{p=1}^{\infty} c_{p}\frac{q^{p}}{ p!},
\end{equation}
where the $c_p \in \mathbb{N}$ coefficients are known as log-cumulants. This coefficients synthesize relevant information of the function and their computation replaces the  estimation of $\zeta(q)$. Computation of log-cumulants is done through the linear regression given by:

\begin{equation}\label{formula_c}
	\hat{c}_m = (\log_2{e}) \displaystyle \sum_{j=j_{min}}^{j_{max}} w_j\hat{C}_m(j). 
\end{equation}

In \citep{wendt2009wavelet}, the computation of $\mathcal{L}(q)$ and $h(q)$ is proposed in its parametric form:

\begin{equation}\label{formula_h}
	\hat{h}(q)= \displaystyle \sum_{j=j_{min}}^{j_{max}} w_jV(2^j, q)
\end{equation}
\begin{equation}\label{formula_D}
	\hat{\mathcal{L}}(q)=\displaystyle \sum_{j=j_{min}}^{j_{max}} w_jV(2^j, q),
\end{equation}
where:

\begin{equation}
	V(j, q) = \frac{\displaystyle \sum_{k=1}^{n_j} L ^q_X (j, k) \log_2{L_X(j, k)}}{n_jS_L(j, q)},
\end{equation} 
\begin{equation}
	U(j, q) = \frac{\displaystyle \sum_{k=1}^{n_j} L ^q_X (j, k) (\log_2{L_X(j, k)} - \log_2{n_j S_L(j, q)})}{n_jS_L(j, q)} + \log_2{n_j}.
\end{equation} 

In this work, equations (\ref{formula_c}), (\ref{formula_h}) and (\ref{formula_D}) are used to compute multifractal features through the computational methods described in the following sections.

	\section{Methods and Experiments}

The objective of this paper is to describe an experiment of lung nodule detection in chest X-Rays, using wavelet-leader based multifractal analysis for feature extraction. The experiment includes three main steps: (i) data selection and preprocessing, (ii) feature selection and extraction, (iii) classification. A detailed description of the steps is given below.

\subsection{Data Selection and Preprocessing}

The database used for the purposes of this work was created by the Digital Images Database Project Team of the Japanese Society of Radiology, within the period from April, 1995 to March, 1997, \citep{shiraishi2000development}. This database contains 247 images of posteroanterior digital chest X-rays, in \lstinline|IMG| format, 154 belong to lungs with nodules and the remaining 93 without nodules. 

Due to the complexity of chest X-rays, because of the variety of anatomical structures and the presence of noise, preprocessing is an important step to achieve better detection performance. To enhance image quality, two techniques were used: histogram equalization, and morphological operations combined with discrete wavelet transform. 

The last preprocessing step is automatic segmentation that, for the purposes of this work, was simplified by using previously computed masks found in an open-source database by \citep{lungsegmentationdk}. Segmentation was performed by multiplying each binary mask to the matrix representing the original image, and then a Gaussian filter for edge smoothing was applied so the multifractal analysis algorithm could preserve robustness when performing over the edge area.

\subsection{Feature Extraction}

Three feature groups were used:

\begin{enumerate}
	\item \textbf{Multifractal features:} Regularity coefficient $h(q)$ and multifractal spectrum $D(q)$ for $q \in \left[0, 2\right]$ and log-cumulants $c_1, c_2$ and $c_3$. To compute those magnitudes wavelet leader based multifractal formalism was used.
	\item \textbf{Other classical texture features:} correlation, second angular moment, homogeneity, contrast, dissimilarity and energy, \citep{texturefeatures}.
	\item \textbf{A combination of both}. 
	
\end{enumerate}

\subsubsection{Feature Preprocessing}

Scaling of numerical values is critical when using a classification algorithm. For example, Support Vector Machine uses dot product in kernel functions; so extremely high values can lead to numerical instability, \citep{hsu2003practical}; moreover, it may cause some features to be dominant over others and, consequently, the algorithm do not learn properly.

\begin{figure}[ht]
	\centering
	\resizebox*{12cm}{!}{\includegraphics{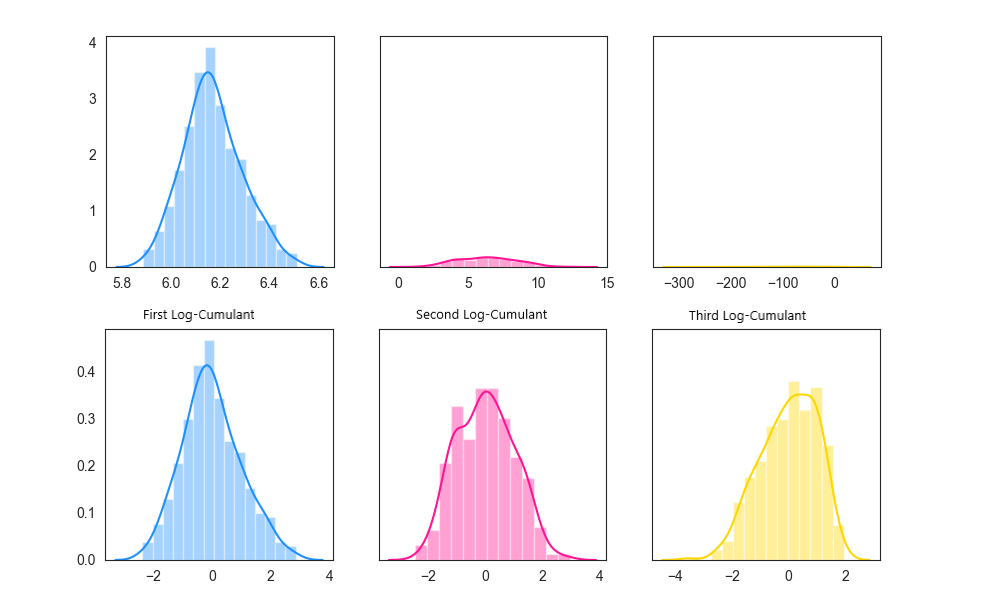}}
	\caption{Histogram of the log-cumulants $c_1$, $c_2$ and $c_3$ before (up) and after (down) scaling.} \label{cumulants}
\end{figure}

\begin{figure}[ht]
	\centering
	\resizebox*{12cm}{!}{\includegraphics{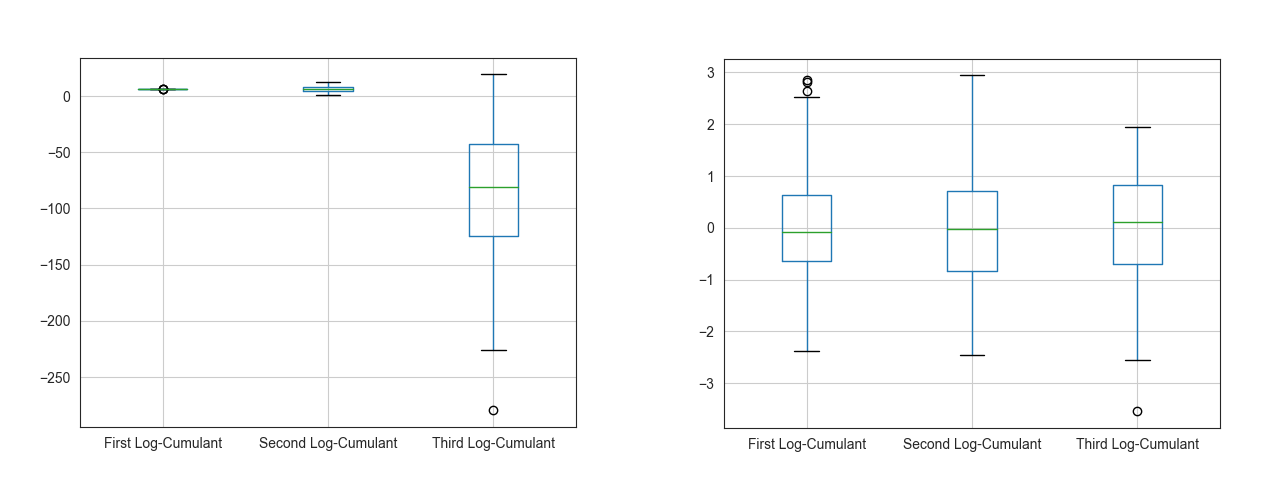}}
	\caption{$c_1$, $c_2$ and $c_3$ log-cumulants box-plot before and after scaling. } \label{boxplot}
\end{figure}

Unbalanced distribution of values can be observed in log-cumulants (see Fig. \ref{cumulants}), whereas values from the first log-cumulant follow a distribution $N(6,0)$, values from the second log-cumulant have a similar average but are more sparse with a standard deviation of $\sigma = 8$. Values from the third log-cumulant are highly sparse, and its average is very far from $c_1$ and $c_2$. In Fig. \ref{boxplot}, dominance of $c_3$ can be observed. After scaling, homogeneity is fairly achieved.


\subsection{Classification Experiment}

Support Vector Machine (SVM) algorithm is a classical binary classifier that was used to perform nodule detection task, \citep{svminproceedings}. Dataset was splitted by using $k-$folding with cross validation to get more reliable results. Other parameters are optimized to fit model to the nature of the dataset. 

\subsubsection{Parameter Optimization}

One of the meta-problems to solve in classification tasks is the optimization of parameters, in order to fit them to the case of study. In this work, \lstinline|GridSearch| class is used for that purpose, \citep{gridsearch}. Its implementation can be found in \lstinline|sklearn.model_selection|, \citep{scikit-learn}.

The following parameters were optimized:
\begin{itemize}
	\item \lstinline|kernel|: Polynomic, Radial Basis Function (RBF) and linear kernels were explored. 
	\item \lstinline|C|: This parameter controls trade-off between misclassified instances and margin width. Set $C \in \{ 2^i | i \in \left[-1, 5 \right] \}$ was explored. 
	\item \lstinline|class_weight|: This parameter is related to unbalanced datasets and it is used to work together with $C$:
	\begin{equation}
		C_i=\omega_i*C.
	\end{equation}
	\lstinline|1:1| or (equal weighting) and \lstinline|'balanced'| proportions were used. In the first case $C_i = C \;\; \forall i$; in the second case $C_i=\frac{1}{n_i}*C$, where $n_i$ is the relative frequency of $ith$ class. 
\end{itemize}

The quality of classification process will be measured with precision, recall and F-score metrics, \citep{metrics}. 

 \subsection{Data Augmentation}
To improve data performance, data augmentation technique was applied. Three new sets of 32 images were obtained from the original dataset, by carefully performing operations of shearing with a scale of 0.02, zooming with a scale of 0.02, and rotation with an angle of $5\deg$; so that anatomical structure of X-rays was not violated.

\subsection{Multifractal Analysis Tool}

Multifractal analysis tool was developed as a \lstinline|Python| module that represents the core of the feature extraction process. 

The main class of the module is \lstinline|Multifractal_Analyser|, that contains the logic involving multifractal features estimation. It is initialized with the following parameters:

\begin{itemize}
	\item \lstinline|min_scale|, \lstinline|max_scale|: Minimum and maximum scales, respectively, default values are \lstinline|2| and \lstinline|5| . 
	\item \lstinline|q|: List that contains all related coefficients, linked to each statistical moment. Default value is \lstinline|[2]|.
	\item \lstinline|wavelet_name|: Mother wavelet used. Default value is \lstinline|db3| (Daubechies with three vanishing moments).
	\item \lstinline|normalization|: Normalization that will be used for leader coefficients, default value is \lstinline|2|, indicating $L^2$ normalization.
	\item \lstinline|gamint|: Fractional integration parameter. It is selected on the basis of $h_{min}+ \eta > 0$.
	\item \lstinline|n_cumulants|: Number of log-cumulants. Default value is \lstinline|3|.
\end{itemize}

The only step needed to trigger multifractal analysis is the invocation of this class\footnote{Default configuration is used}: \lstinline|mfa= MultifactalAnalyser; mfa()|, as the whole logic is implemented in the \underline{\hspace{0.5cm}}\lstinline|call|\underline{\hspace{0.5cm}} method.

As an auxiliary class,  \lstinline|MultiresolutionQuantity| is implemented. It express the logic of multiresolution quantities by storing a dictionary of the $scale \rightarrow values$ correspondence.  

For the estimation of multifractal spectrum, \lstinline|wavelet_coeff| and \lstinline|wavelet_leaders| were implemented, to store wavelet and leader coefficients respectively. 

Computation of structure function $S_L(q, j)$, log-cumulants $c_m$ and \textit{Legendre} spectrum $\mathcal{L}$ was done by the method presented in \ref{practical_mfa}.

\section{Results and Discussion} 

\begin{figure}[ht]
	\centering
	\resizebox*{12cm}{!}{\includegraphics{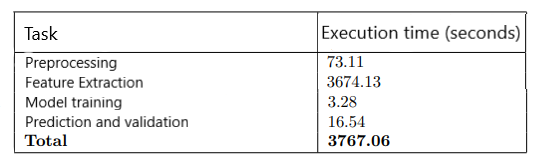}}
	\caption{Computational cost by general task.} \label{comp_cost}
\end{figure}

In this section, main results are presented and discussed. In each step of the experiment an in-depth analysis of every related variable was made, in terms of computational cost and model performance (See Fig. \ref{comp_cost}). 

\begin{figure}[ht]
	\centering
	\resizebox*{15cm}{!}{\includegraphics{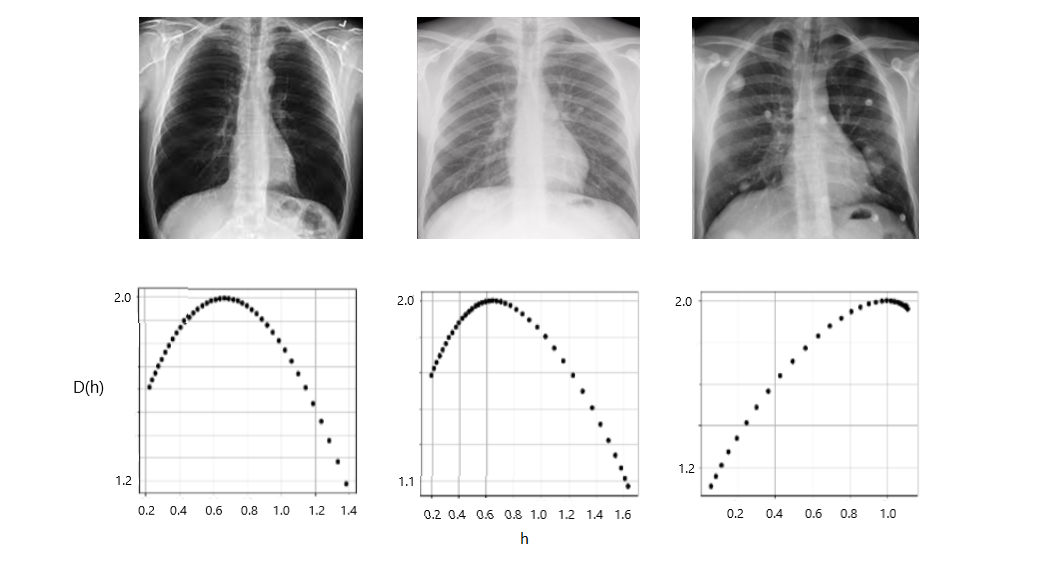}}
	\caption{Multifractal spectrum $\left\langle h, D(h)\right\rangle $ of three chest X-rays: two healthy lungs images (left) and lungs with nodules (right).} \label{visual_inspection}
\end{figure}

To set a starting point for the validity of the proposed method, a set of 20 X-rays was inspected. In the case of healthy lungs, multifractal spectrum shows a maximum for a singularity exponent between $0.6$ and $0.8$, with symmetrical behavior. However, in lungs with nodules, multifractal spectrum has its maximum for a singularity exponent near 1 and it is not symmetrical. Lung area in chest X-rays can be visually erratic and this can lead to the presence of isolated singularities in healthy lungs, what would explain the high rate of values with a \textit{H\"older} exponent less than 1. However, lung nodules are dense and compact structures; that is why, within intervals contained in the structure of a nodule, a more steady $h$ exponent is found, to denote smoothness and leading to $D(h)$ maximum to be reached at values near to 1 (See Fig. \ref{visual_inspection}).

\begin{figure}[ht]
	\centering
	\resizebox*{10cm}{!}{\includegraphics{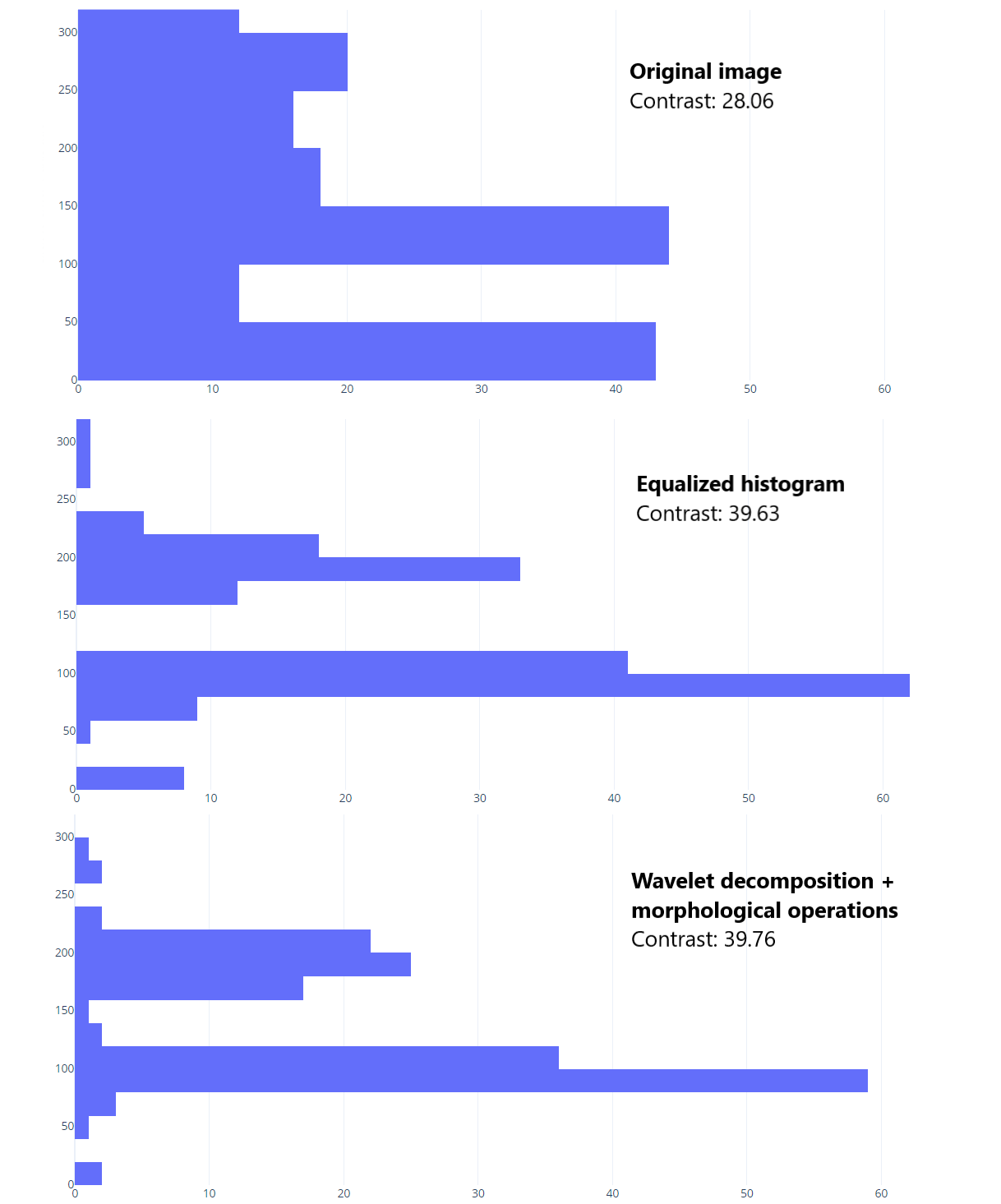}}
	\caption{Histograms of a lung X-Ray.} \label{hist_contrast}
\end{figure}

\begin{figure}[ht]
	\centering
	\resizebox*{12cm}{!}{\includegraphics{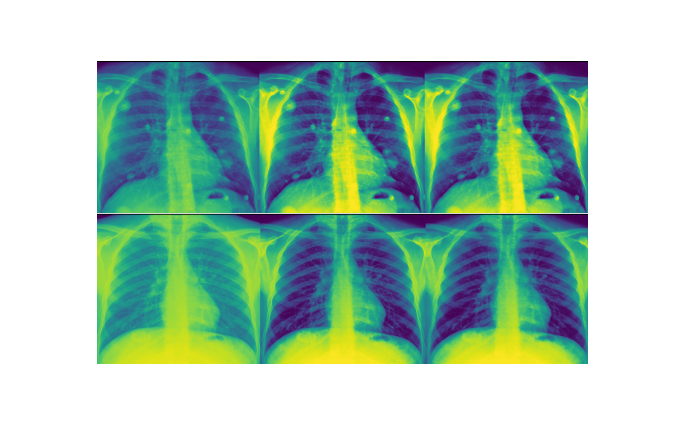}}
	\caption{Enhancement of two X-rays: lungs with nodules (top), healthy lungs (bottom); from left to right: original, enhancement with histogram equalization, enhancement with wavelet decomposition and morphological operations of opening and closure, \citep{aidoo2019chest}.} \label{enhancement}
\end{figure}

In the image preprocessing step, histogram equalization proved to substantially improve image visualization; however, a combination of wavelet transform and morphological operations shows a slight blurring of lung opacities.  This could appear as a limitation to some applications of contrast enhancement, but in the case of nodule detection it allows the classification algorithm to ignore other opacities and singularities different from nodules that, because of its compactness and well-defined borders, remain observable after applying method in Fig. \ref{enhancement}. In fact, contrast measure was slightly better after applying the latter method (See Fig. \ref{hist_contrast}). 

\begin{figure}[ht]
	\centering
	\resizebox*{12cm}{!}{\includegraphics{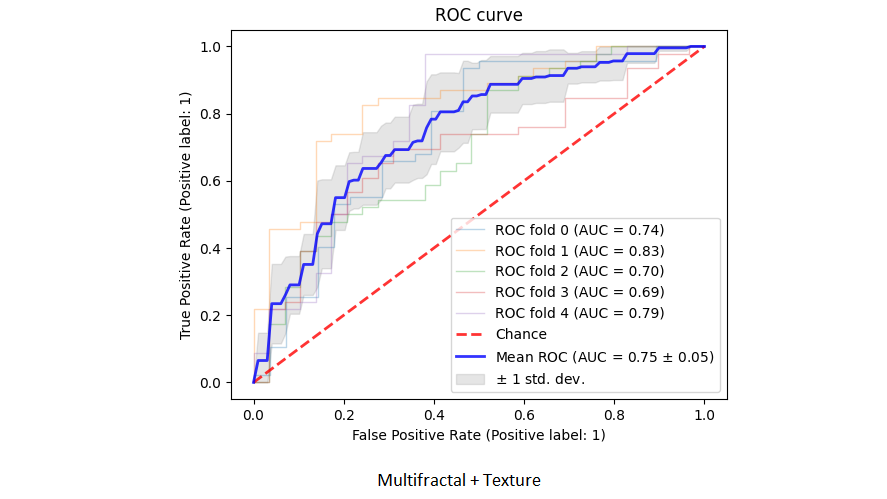}}
	\caption{ROC curve for detection task using multifractal features combined with texture features.} \label{roc_auc}
\end{figure}

\begin{figure}
	\centering
	\subfloat[ROC curve obtained when using multifractal features only.]{%
		\resizebox*{7cm}{!}{\includegraphics{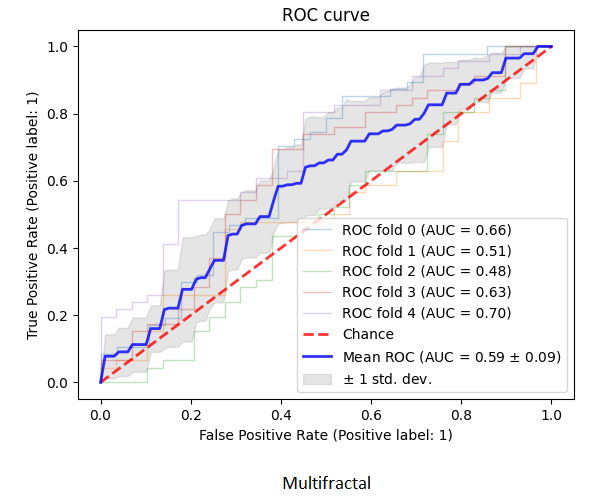}}}\hspace{5pt}
	\subfloat[ROC curve obtained when using classical texture features only.]{%
		\resizebox*{7cm}{!}{\includegraphics{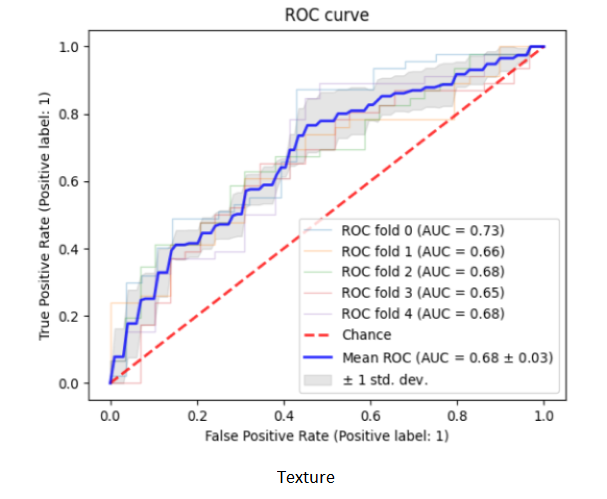}}} 
	\caption{Comparison of ROC curves obtained after using two different sets of features. } \label{roc_comparison}
\end{figure}

After exploring different combinations of values for parameters $C$, class weighting and kernel,  through grid-search tool and cross validation, classification task yielded a best result of 75\% precision, 75\% recall and $75\%$ $F_\beta$ score (See Fig. \ref{roc_auc}). Regarding feature selection, those results where obtained with singularity exponent $h$; multifractal spectrum $D(q)$; log-cumulants $c_1, c_2, c_3$; correlation and contrast (See Figs. \ref{roc_auc}, \ref{roc_comparison}). This results suggest that classification is more effective when multifractal descriptors are not used alone. 

\begin{figure}[ht]
	\centering
	\resizebox*{14cm}{!}{\includegraphics{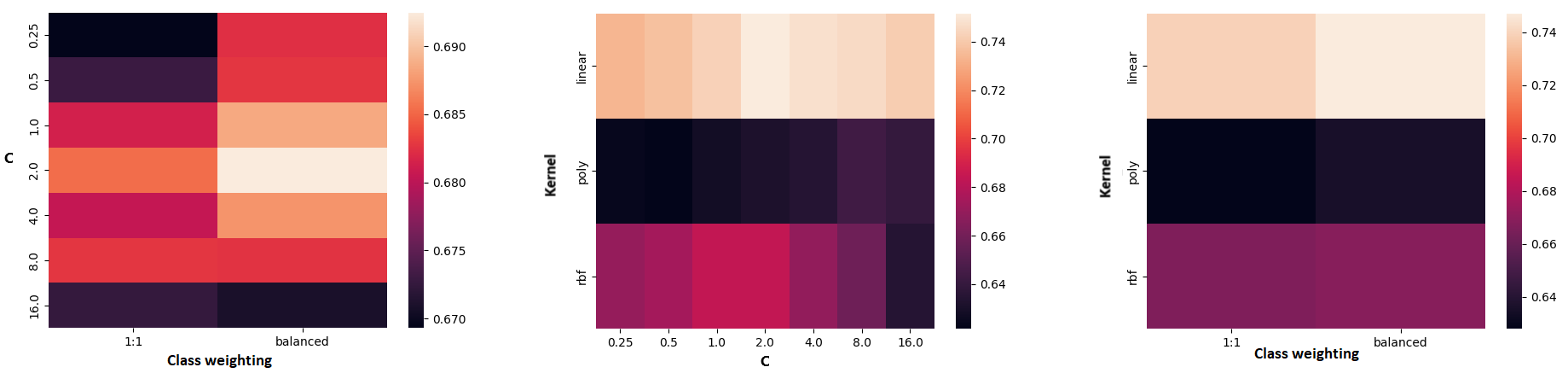}}
	\caption{Grid search results for the three possible combination of parameters. } \label{grid_search}
\end{figure}

Regarding parameter optimization for SVM, best results were obtained when using a linear kernel. A constant $C=2$ and a balanced class weighting were also found to be the best fit to the problem (See Fig. \ref{grid_search}).

Data augmentation technique slightly improved ROC AUC metric in $\approx 0.04$. In order to avoid overfitting, the amount of generated images was selected after analyzing the learning curves of both the validation and training sets. 

\section{Conclussions}
We present the results of an investigation applying multifractal analysis for detection of lung nodules in X-ray images, based on their texture. The present study also compares descriptors of the multifractal spectrum against and in conjunction with texture features like correlation, contrast, among others.

Regarding texture analysis of the lung nodules, the experiments show that multifractal information can provide better results when combined with other classical texture features. The best performance was achieved combining the singularity exponent, multifractal spectrum and log-cumulants with correlation and contrast. 

In classification task, Support Vector Machine algorithm was used. The experiments included an optimization of the parameters in the learning algorithm, providing best results for a linear kernel, a constant $C=2$ and a balanced class weighting. The best ROC AUC performance registered was of $0.75 \pm 5$. Between precision and recall, recall was seen as more relevant, taking into consideration that in medical imaging it is more critical to detect most positive cases than to discard negative ones, according to the professionals consulting this research. That is why in $F_\beta$ metric a value of $\beta = 2$ was used. A value of $F_\beta = 0.75$ was obtained. 

\begin{figure}[ht]
	\centering
	\resizebox*{8cm}{!}{\includegraphics{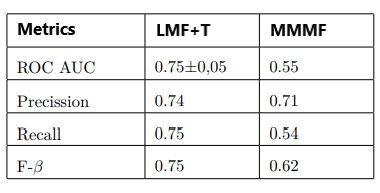}}
	\caption{Comparison between proposed method (referenced as \textbf{LMF+T}) and Modulus Maxima Wavelet-based in \citep{juanda} (referenced as \textbf{MMMF}). } \label{mma_comp}
\end{figure}

Multifractal features estimation was based on the wavelet leader formalism. Results were compared with the estimation using Modulus Maxima wavelet transform based formalism \citep{juanda}, the former being better with an overall out-performance of approximately 20 \% (See Fig. \ref{mma_comp}).  

The present study also has demonstrated that the wavelet leader coefficients meet all the requirements to be considered as a multirresolution quantity, namely: 
\begin{enumerate}[i]
	\item to be a hierarchical quantity,
	\item to be robust to smooth trends, achieved through vanishing moments of mother wavelet
	\item to have a decay described by power law with $\alpha = h$, guaranteed through linear regression.
\end{enumerate}

Our proposal also includes a preprocessing module, where two contrast enhancement techniques, histogram equalization and morphological operations over wavelet reconstruction, were used. Enhancement is followed by two segmentation steps: first, the lung area is separated from the X-Ray using a mask and edge smoothing techniques were applied to achieve a more robust image analysis.

The highest temporal cost was reached at feature extraction, taking 97\% of total computing time.  Multifractal analysis tool has a high temporal cost; however, its operations are easy to implement. In this work, data structures were used to attenuate the first.

\bibliographystyle{unsrtnat}
\bibliography{references}

\begin{thebibliography}{32}
\providecommand{\natexlab}[1]{#1}
\providecommand{\url}[1]{\texttt{#1}}
\expandafter\ifx\csname urlstyle\endcsname\relax
  \providecommand{\doi}[1]{doi: #1}\else
  \providecommand{\doi}{doi: \begingroup \urlstyle{rm}\Url}\fi

\bibitem[Kelly(2012)]{kelly2012chest}
Barry Kelly.
\newblock The chest radiograph.
\newblock \emph{The Ulster medical journal}, 81\penalty0 (3):\penalty0 143,
  2012.

\bibitem[Seuret(2016)]{seuret2016multifractal}
St{\'e}phane Seuret.
\newblock Multifractal analysis and wavelets.
\newblock In \emph{New Trends in Applied Harmonic Analysis}, pages 19--65.
  Springer, 2016.

\bibitem[Kido et~al.(1995)Kido, Ikezoe, Naito, Tamura, and
  Machi]{kido1995fractal}
Shoji Kido, Junpei Ikezoe, Hiroaki Naito, Shinichi Tamura, and Setuko Machi.
\newblock Fractal analysis of interstitial lung abnormalities in chest
  radiography.
\newblock \emph{Radiographics}, 15\penalty0 (6):\penalty0 1457--1464, 1995.

\bibitem[Rodr{\'\i}guez et~al.(2005)Rodr{\'\i}guez, Lemus, Serrano, Casadiego,
  and Correa]{rodriguez2005medidas}
Javier Rodr{\'\i}guez, Jorge Lemus, Julio Serrano, Elkin Casadiego, and Calina
  Correa.
\newblock Medidas fractales cardiotor{\'a}cicas en radiograf{\'\i}as de
  t{\'o}rax.
\newblock \emph{Revista Colombiana de Cardiología}, 12\penalty0 (3):\penalty0
  129--134, 2005.

\bibitem[Silvetti and Delrieux(2010)]{silvetti2010analisis}
Andrea Silvetti and Claudio Delrieux.
\newblock An{\'a}lisis multifractal aplicado a im{\'a}genes m{\'e}dicas.
\newblock In \emph{XII Workshop de Investigadores en Ciencias de la
  Computaci{\'o}n}, Patagonia, Argentina, 2010.

\bibitem[Braverman and Tambasco(2013)]{braverman2013scale}
Boris Braverman and Mauro Tambasco.
\newblock Scale-specific multifractal medical image analysis.
\newblock \emph{Computational and Mathematical Methods in Medicine}, 2013,
  2013.
\newblock \doi{10.1155/2013/262931}.

\bibitem[Maceda~Macario(2014)]{maceda2014computo}
Violeta Maceda~Macario.
\newblock C{\'o}mputo de medidas fractales para im{\'a}genes m{\'e}dicas.
\newblock Master's thesis, Benem{\'e}rita Universidad Aut{\'o}noma de Puebla,
  2014.

\bibitem[Leonarduzzi(2014)]{leonarduzzi2014analisis}
Roberto~Fabio Leonarduzzi.
\newblock \emph{An{\'a}lisis multifractal basado en coeficientes ondita
  l{\'\i}deres: formalismo multifractal basado en p-l{\'\i}deres y
  aplicaci{\'o}n a se{\~n}ales biom{\'e}dicas}.
\newblock PhD thesis, 2014.

\bibitem[Hern{\'a}ndez and C{\'o}rdova~Fraga(2015)]{hernandez2015analisis}
Jessica Janett~{\'A}vila Hern{\'a}ndez and Teodoro C{\'o}rdova~Fraga.
\newblock An{\'a}lisis de im{\'a}genes m{\'e}dicas usando matlab.
\newblock \emph{J{\'o}venes en la Ciencia}, 1\penalty0 (2), 2015.

\bibitem[Troshin(2015)]{troshin2015multifractal}
PI~Troshin.
\newblock Multifractal parametrization in diagnosis of lungs diseases.
\newblock \emph{International Journal of Pure and Applied Mathematics},
  105\penalty0 (2):\penalty0 173--185, 2015.

\bibitem[Marusina et~al.(2017)Marusina, Mochalina, Frolova, Satikov, Barchuk,
  Kuznetcov, Gaidukov, and Tarakanov]{marusina2017mri}
Mariya~Y Marusina, Alexandra~P Mochalina, Ekaterina~P Frolova, Valentin~I
  Satikov, Anton~A Barchuk, Vladimir~I Kuznetcov, Vadim~S Gaidukov, and
  Segrey~A Tarakanov.
\newblock Mri image processing based on fractal analysis.
\newblock \emph{Asian Pacific Journal of Cancer Prevention (APJCP)},
  18\penalty0 (1):\penalty0 51--55, 2017.

\bibitem[León~Arencibia(2019)]{lorenaleon}
Lorena León~Arencibia.
\newblock Clasificación de malignidad de masas en mamografías digitales a
  través del análisis multifractal.
\newblock Master's thesis, Universidad de La Habana, 2019.

\bibitem[Lee and Jwo(2020)]{lee2020automatic}
Cheng-Hsiung Lee and Jung-Sing Jwo.
\newblock Automatic segmentation for pulmonary nodules in ct images based on
  multifractal analysis.
\newblock \emph{IET Image Processing}, 14\penalty0 (7):\penalty0 1347--1353,
  2020.

\bibitem[Xu et~al.(2002)Xu, Ahuja, and Bansal]{xu2002automated}
Ning Xu, Narendra Ahuja, and Ravi Bansal.
\newblock Automated lung nodule segmentation using dynamic programming and
  em-based classification.
\newblock In \emph{Medical Imaging 2002: Image Processing}, volume 4684, pages
  666--676. International Society for Optics and Photonics, 2002.

\bibitem[Kostis et~al.(2003)Kostis, Reeves, Yankelevitz, and
  Henschke]{kostis2003three}
William~J Kostis, Anthony~P Reeves, David~F Yankelevitz, and Claudia~I
  Henschke.
\newblock Three-dimensional segmentation and growth-rate estimation of small
  pulmonary nodules in helical ct images.
\newblock \emph{IEEE transactions on medical imaging}, 22\penalty0
  (10):\penalty0 1259--1274, 2003.

\bibitem[Mandelbrot and Mandelbrot(1982)]{mandelbrot1982fractal}
Benoit~B Mandelbrot and Benoit~B Mandelbrot.
\newblock \emph{The fractal geometry of nature}, volume~1.
\newblock WH Freeman New York, 1982.

\bibitem[Liebovitch and Toth(1989)]{LIEBOVITCH1989386}
Larry~S. Liebovitch and Tibor Toth.
\newblock A fast algorithm to determine fractal dimensions by box counting.
\newblock \emph{Physics Letters A}, 141\penalty0 (8):\penalty0 386--390, 1989.
\newblock ISSN 0375-9601.
\newblock \doi{https://doi.org/10.1016/0375-9601(89)90854-2}.
\newblock URL
  \url{https://www.sciencedirect.com/science/article/pii/0375960189908542}.

\bibitem[Salat et~al.(2016)Salat, Murcio, and Arcaute]{multifractalspectrum}
Hadrien Salat, Roberto Murcio, and Elsa Arcaute.
\newblock Multifractal methodology.
\newblock \emph{Physica A: Statistical Mechanics and its Applications}, 473, 06
  2016.
\newblock \doi{10.1016/j.physa.2017.01.041}.

\bibitem[Sisworo(1999)]{article}
Sisworo Sisworo.
\newblock On holder exponents.
\newblock \emph{Matematika \& Sains}, 4:\penalty0 244--249, 01 1999.

\bibitem[Wendt et~al.(2009)Wendt, Roux, Jaffard, and Abry]{wendt2009wavelet}
Herwig Wendt, St{\'e}phane~G Roux, St{\'e}phane Jaffard, and Patrice Abry.
\newblock Wavelet leaders and bootstrap for multifractal analysis of images.
\newblock \emph{Signal Processing}, 89\penalty0 (6):\penalty0 1100--1114, 2009.

\bibitem[Mallat(2008)]{mallat1999wavelet}
St{\'e}phane Mallat.
\newblock \emph{A wavelet tour of signal processing: the sparse way}.
\newblock Elsevier, tercera edition, 2008.

\bibitem[Castaing et~al.(1993)Castaing, Gagne, and Marchand]{castaing1993log}
Bernard Castaing, Yves Gagne, and Muriel Marchand.
\newblock Log-similarity for turbulent flows?
\newblock \emph{Physica D: Nonlinear Phenomena}, 68\penalty0 (3-4):\penalty0
  387--400, 1993.

\bibitem[Shiraishi et~al.(2000)Shiraishi, Katsuragawa, Ikezoe, Matsumoto,
  Kobayashi, Komatsu, Matsui, Fujita, Kodera, and
  Doi]{shiraishi2000development}
Junji Shiraishi, Shigehiko Katsuragawa, Junpei Ikezoe, Tsuneo Matsumoto,
  Takeshi Kobayashi, Ken-ichi Komatsu, Mitate Matsui, Hiroshi Fujita, Yoshie
  Kodera, and Kunio Doi.
\newblock Development of a digital image database for chest radiographs with
  and without a lung nodule: receiver operating characteristic analysis of
  radiologists' detection of pulmonary nodules.
\newblock \emph{American Journal of Roentgenology}, 174\penalty0 (1):\penalty0
  71--74, 2000.

\bibitem[Rusak et~al.(2019)Rusak, Wang, and Arzhaeva]{lungsegmentationdk}
Filip Rusak, Dadong Wang, and Yulia Arzhaeva.
\newblock Lung segmentation data kit. v1. csiro. data collection, 2019.
\newblock URL \url{https://doi.org/10.25919/5c49548be0551}.

\bibitem[Haralick et~al.(1973)Haralick, Shanmugam, and
  Dinstein]{texturefeatures}
Robert Haralick, Kalaivani Shanmugam, and IH~Dinstein.
\newblock Texture features for image classification.
\newblock \emph{IEEE Transactions on Systems, Man, and Cybernetics},
  SMC-3\penalty0 (6):\penalty0 610--621, 01 1973.
\newblock \doi{10.1109/TSMC.1973.4309314}.

\bibitem[Hsu et~al.(2003)Hsu, Chang, and Lin]{hsu2003practical}
Chih-Wei Hsu, Chih-Chung Chang, and Chih-Jen Lin.
\newblock A practical guide to support vector classification, 01 2003.

\bibitem[Evgeniou and Pontil(2001)]{svminproceedings}
Theodoros Evgeniou and Massimiliano Pontil.
\newblock Support vector machines: Theory and applications.
\newblock volume 2049, pages 249--257, 01 2001.
\newblock \doi{10.1007/3-540-44673-7_12}.

\bibitem[Brownlee(2020)]{gridsearch}
Jason Brownlee.
\newblock Hyperparameter optimization with random search and grid search.
\newblock \emph{Machine Learning Mastery}, 09 2020.
\newblock URL
  \url{https://machinelearningmastery.com/hyperparameter-optimization-with-random-search-and-grid-search/}.
\newblock Accessed: 2022-01-23.

\bibitem[Pedregosa et~al.(2011)Pedregosa, Varoquaux, Gramfort, Michel, Thirion,
  Grisel, Blondel, Prettenhofer, Weiss, Dubourg, Vanderplas, Passos,
  Cournapeau, Brucher, Perrot, and Duchesnay]{scikit-learn}
F.~Pedregosa, G.~Varoquaux, A.~Gramfort, V.~Michel, B.~Thirion, O.~Grisel,
  M.~Blondel, P.~Prettenhofer, R.~Weiss, V.~Dubourg, J.~Vanderplas, A.~Passos,
  D.~Cournapeau, M.~Brucher, M.~Perrot, and E.~Duchesnay.
\newblock Scikit-learn: Machine learning in {P}ython.
\newblock \emph{Journal of Machine Learning Research}, 12:\penalty0 2825--2830,
  2011.

\bibitem[Goyal(2021)]{metrics}
Shweta Goyal.
\newblock Evaluation metrics for classification models.
\newblock \emph{Analytics Vidhya}, 07 2021.
\newblock URL
  \url{https://medium.com/analytics-vidhya/evaluation-metrics-for-classification-models-e2f0d8009d69}.
\newblock Accessed: 2021-12-06.

\bibitem[Aidoo et~al.(2019)Aidoo, Wilson, and Botchway]{aidoo2019chest}
Anthony~Y Aidoo, Matilda Wilson, and Gloria~A Botchway.
\newblock Chest radiograph image enhancement with wavelet decomposition and
  morphological operations.
\newblock \emph{TELKOMNIKA Telecommunication Computing Electronics and
  Control}, 17\penalty0 (5):\penalty0 2587--2594, 2019.

\bibitem[Menéndez~del Cueto(2021)]{juanda}
Juan~David Menéndez~del Cueto.
\newblock Detección de nódulos en radiografías de torax a través del
  formalismo multifractal basado en la transformada wavelet de módulo máximo,
  2021.
\newblock Bachelor thesis, Universidad de La Habana.

\end{thebibliography}

\end{document}